# Optical Braille Recognition Using Object Detection CNN

Ilya G. Ovodov
ELVEES RnD center, JSC
Zelenograd, Russia
il@ovdv.ru

*Abstract*—This paper proposes an optical Braille recognition method that uses an object detection convolutional neural network to detect whole Braille characters at once. The proposed algorithm is robust to the deformation of the page shown in the image and perspective distortions. It makes it usable for recognition of Braille texts being shoot on a smartphone camera, including bowed pages and perspective distorted images. The proposed algorithm shows high performance and accuracy compared to existing methods. We also introduce a new "Angelina Braille Images Dataset" containing 240 annotated photos of Braille texts. The proposed algorithm and dataset are available at GitHub.

I   INTRODUCTION

The embossed Braille alphabet was invented in 1824 and served as the primary way of writing and reading for blind people for many years. Recent advances in technology provide blind people with many new opportunities for receiving and transmitting the information. However, reading printed Braille and writing in Braille continues to be an important communication method for them. However, Braille is often used for communication between blind and sighted people. In particular, the situation is quite common when sighted teachers work with blind students, and they have to deal with textbooks and student works written in Braille.

Braille is an alphabet made of dots embossed on paper for tactile recognition. Nevertheless, sighted people who deal with Braille texts usually do not have enough tactile skill to read the text with their fingers and are forced to read it with their eyes. However, braille text is visually represented by white raised points on a white background, so its' visual recognition is extremely tedious. Reading double-sided Braille is especially difficult. Text printed on the back of the paper looks like dented dots. Tactically, they are practically invisible and do not interfere with the sensation of the convex points of the front side, but visually they are hardly distinguishable from them (Fig.1), so reading such a text with the eyes is especially difficult. The use of technical means, in particular, optical recognition methods, for Braille recognition, can greatly facilitate this work. Therefore, optical braille recognition methods have been developing since at least the 1980s [1].

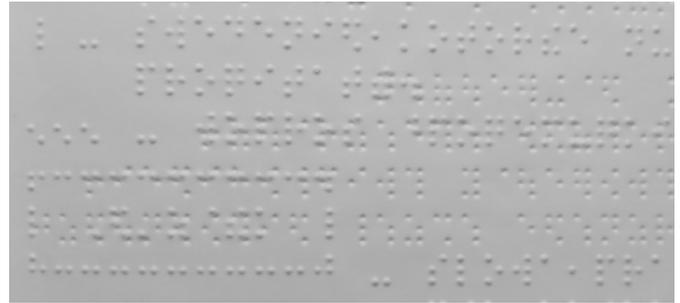

Fig.1   Example of a page Taken with double-sided printing

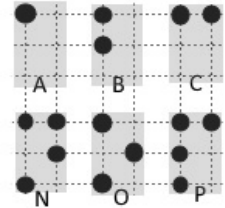

Fig.2   Schematic arrangement of Braille text

Each letter or other character is encoded in a Braille text with several (1 to 6) bulging points located in a 2x3 grid, allowing 63 different characters to be encoded. The distance between adjacent symbols is slightly larger than between two columns of points in one symbol. The character width, the spacing between characters, the spacing between lines, and the characters' places on a line are constant for each Braille document. Thus, the dots in the Braille text are attached to a fixed grid (Fig.2). The use of this fact is critical to most existing methods of Braille recognition. However, this significantly limits their applicability. These algorithms either require a scanner or special photographing conditions to ensure the Braille page's image accurate alignment.

This work aims to provide the recognition method applicable to Braille text images obtained on a mobile phone camera in the domestic environment. The grid Braille characters are attached to can be significantly distorted due to: a) perspective distortions caused by the fact that the sheet is not perpendicular to the camera optical axis, b) the paper sheet curvature on open book spread images. Also, different areas of the sheet can have significantly different lighting (Fig.3). For this purpose, we use an approach based on the use of object detection convolutional neural networks.

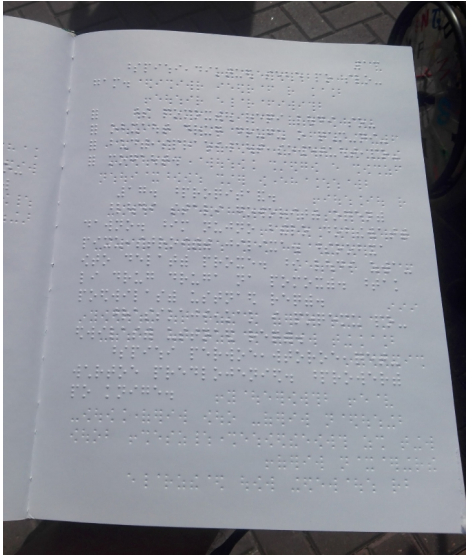

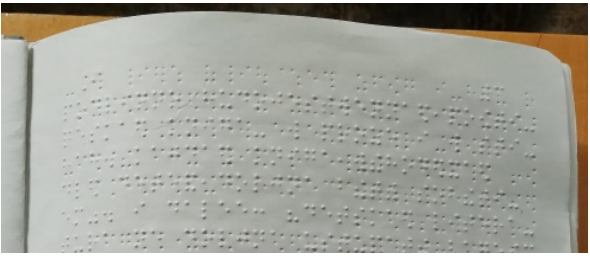

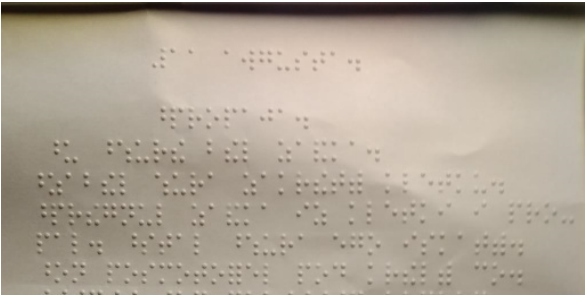

Fig.3 Examples of Braille text smartphone photos with perspective distortion (a), page curvature (b) and deformation (c).

## II Related work

### A Optical braille recognition approaches

The main approach to optical braille recognition consists of sequentially performing Braille points search, characters grid restoration, including compensation for possible image rotation, points grouping into characters, and, finally, character decoding. The survey papers [2], [3] consider all algorithms in accordance with this sequence of steps.

The simplest approach to point detection is thresholding. Zhang and Yoshino [4] use a dynamic local threshold. In order to detect dots on double-sided braille and distinguish between the front side and reverse side dots, Antonacopoulos and Bridson [5] use detection of light and dark areas of dots, and segment image into areas of three classes: bright, dark, and background. For this, they use static thresholds relative to the average brightness level in the vicinity of the point. Renqiang Li et al. [6] use Haar detector and SVM. Morgavi and Morando [7] use a simple neural network to find points. Venugopal-Wairagade [8] fulfill circle detection using Hough transform. Perera, Wanniarachchi [9] use HOG and SVM, R.Li et al. [6] - Haar detector and Adaboost., R.Li et al. [10] - Haar detector and Adaboost for primary dot detection and HOG, LBP and SVM for final dot detection after grid restoration.

The next step is to restore the grid to which the points are anchored and possibly compensate for the sheet rotation. For this purpose can be used linear regression [9], Hough Transform [5], [11], or coordinates distribution density during step-by-step rotation of the image can be investigated to find optimum rotation angle [10]. Sometimes, after the image de-skewning, points are searched again using information about their possible position [5], [10].

Although some methods assume the grid deformation, implying a change in the pitch between different lines ([10] and others), it is assumed that the grid lines are straight on the entire sheet and parallel.

Since the mentioned works essentially rely on snapping Braille points to the grid, they mostly work with images obtained with a scanner. In this case, the necessary image correction is reduced to de-skewing, which makes the grid lines vertical and horizontal. Only a few works declare the purpose of OBR on images from a smartphone ([4], [8]), but the methods described there are still based on the presence of a rectangular grid on a sheet.

Although convolutional neural networks (CNN) made tremendous advances in image recognition in recent years, the use of deep learning and neural networks for OBR is sparse. There are only a few papers on the use of fully connected neural networks. Morgavi, Morando [7] and Ting Li et al. [12] use a simple neural network to find points, Subur et al. [13] - to find the symbol value using the points found by image segmentation. Kawabe et al. [11] use it for separating front and back points when recognizing two-sided braille. Only R.Li et al. [14] presented at CVPRW 2020 work based on segmentation neural network with modified UNet architecture. They used a neural network to determine areas occupied by Braille characters and recognize these characters. Subsequent post-processing was required to determine the positions of individual symbols based on the segmentation results.

### B Optical braille recognition methods accuracy evaluation

While various works provide different quantitative accuracy characteristics of proposed algorithms, sometimes quite high, comparing algorithms with each other encounters at least two obstacles:

- Until recently, there was no open dataset to compare different algorithms. Quality values provided in papers were measured on proprietary, not published datasets, so it is mostly impossible to compare published results of different works.

- Often works that use the general pipeline described above (i.e., points detection-grid restoration-grouping

points into symbols and decoding), quality indicators are given only for the point recognition stage. It prevents comparing the performance of these algorithms with algorithms that do not have a separate stage of point recognition, such as our algorithm.

The only publicly available DSBI dataset with braille text was published by Li et al. [6]. It contains 114 pages of scanned two-sided braille texts, divided into the train (26 pages) and test (88 pages) sets. All pages are carefully aligned during scanning. A grid of points for the front and back sides has been calculated, rotation required to bring the grid to a vertical-horizontal orientation has been calculated. The annotation is made by specifying the rotation angle, the coordinates of vertical and horizontal grid lines after rotation, and a list of braille characters referenced to this grid's nodes. All texts are in Chinese, but the Braille alphabet has the same structure in all languages, and this dataset can be used regardless of language.

Although this dataset does not look large enough and variable enough to provide full training of recognition algorithms (only 26 pages in the training set), it allows you to compare different approaches to the problem. The authors of [6] compared the accuracy of algorithms based on image segmentation (Antonacopoulos et al., 2004 [5]), Haar features and Adaboost (Viola & Jones [15]), and their algorithm (Li et al. [10]). However, they provide only the accuracy of point detection. Only in [14] they provided accuracy metrics of their algorithm, estimated not at a dot but a character level.

*C  Object detection convolutional neural nets*

The use of convolutional neural networks (CNN) in computer vision and, in particular, in object detection has made tremendous strides in recent years. Convolutional networks were proposed by LeCun 1989 [16], but their popularity has exploded since 2012 [17]. After classification problems, they were applied to solve the object detection problem: the simultaneous finding of rectangular areas containing objects in the image and classifying the objects contained in them. Initially, CNN-based solutions for object detection processed search for regions and classification of objects separately ([18] and others). Later methods that simultaneously search for regions and its classification (one-stage detectors) have achieved great success.

The one-stage detector principle for the object detection has become widespread with the advent of SSD [19], [20] and YOLO [21], [22] convolutional neural network architectures. One-stage detectors' key idea is that after a series of convolutions and size reductions, they produce a feature map, each point of which corresponds to a square region of the original image. Before training, several a priori sizes of desired objects are set for each cell of the feature map, called anchors. For each point of the feature map (i.e., for each square area in the original image), the ground truth annotation boxes intersecting with the anchors are considered, and the degree of intersection is calculated as IOU (intersection over union). The neural network learns to predict the necessary shift and resizing of the anchor (4 outputs per anchor), the degree of intersection (1 output per anchor) which is concerned as confidence at inference stage, indicating presence of an object in the area of anchor, and class of the object (*C* output parameters where *C* is the number of classes). Thus, at each point of the feature map, the neural network learns to predict either *A·(4+1+C)* output values, where *A* is the number of anchors, 4 outputs define coordinates of the bounding box [21], [22] or *A·(4+C)* output values [19], [20], [23]. In the latter case, confidence is included in the responses for all *C* classes, so that for anchors whose areas do not correspond to objects, the responses for all *C* classes are small.

Simultaneous learning of location and class prediction is achieved using the loss function

$$L = L_{loc} + \lambda_{cls} L_{cls} \qquad (1)$$

where $L_{loc}$ is the loss function for location prediction errors, $L_{cls}$ is the loss function for classification errors, $\lambda_{cls}$ is the weighting factor.

The RetinaNet [23] further develops this approach proposing an improved loss function FocalLoss for $L_{cls}$ component of the loss function, which provides more weight for more complex cases. See [23] for details. We used this implementation of object detection CNN in this work.

### III  OUR APPROACH

*A  The problem setting and the network architecture*

Unlike conventional approaches, we do not separate stages of dots detection, grid restoration, and combining dots into characters. Instead, we find whole Braille symbols directly and simultaneously recognize them, using the object detection CNN described above. We assign each character a class label from 1 to 63 using formula $c = \sum_{1}^{6} 2^{i-1} p_i$ where $p_i$ is 1 if the i-th point presents in the character and 0 otherwise.

We scale input image to 100dpi resolution which results in scaling standard A4 page to 864x1150 resolution. Braille characters are located with horizontal spacing of ~25pt and line spacing ~40pt.

We use RetinaNet CNN architecture described in [23] with some minor modifications. Optical Braille Recognition task differs in that it searches for a large number of objects of approximately the same small size and a fixed width to height ratio. Therefore, we have simplified RetinaNet architecture in order to reduce the execution time, primarily NMS operations. Only one "output to class + box subnet" (see Fig.3 in [23]) was used at the layer level with feature map cells having 16x16 size. It guarantees that every Braille character is covered by at least one grid cell. We used only one anchor for each grid cell with a size close to the expected character size. These modifications have reduced calculation time by more than 5 times without substantial loss of recognition quality.

Also, we used an a priory concern that Braille characters do not overlap and substantially reduced the IOU threshold used to filter overlapping detected bounding boxes using the NMS procedure. We used IOU threshold = 0.02 to allow only small overlapping due to detector imperfection.

*B  Angelina Braille Images Dataset*

DSBI dataset [6] was the only publicly available dataset with labeled Braille text images, and we would like to express our deep appreciation to its creators. However, DSBI contains only scanned Braille images where Braille dots are aligned to a rectangular grid. It has rather small number of images with limited diversity. Therefore it is not suitable for training the CNN for the proposed algorithm on images it is designed to handle or for testing it on difficult samples.

We prepared the new "Angelina Braille Images Dataset" containing 212 pages of double-sided braille books and 28 pages of student works. All these texts were photographed with various photo cameras or mobile phones under conditions close to the algorithms' intended work conditions. It includes curved pages in the open spread of the book and perspective distortions. Characters on the front side of the texts are labeled using the usual object detection problem method: for each character, a bounding box is defined, and a class from 1 to 63 is assigned, corresponding to the braille character inside the box. Sample images are shown in Fig.3, sample labels in Fig.4

Annotation of the dataset was produced iteratively. At each iteration:

a) primary annotation was automatically generated;

b) automatically generated annotation bounding boxes were corrected manually to fit Braille text lines;

c) annotation Braille characters were converted into a plain text;

d) poetry texts were checked by comparison with ground truth texts of the same poems found from the Internet, considering that splitting the text into lines for poetry texts is fixed. At later iterations, non-poetry texts were checked by a spell-checker. All questionable cases were checked manually. As a result, when it is not clear if there is a Braille dot at some position or not, all ambiguous cases were resolved in favor of the grammatically correct option.

The fact was noticed that bulging front side dots and bulged-in reverse side dots are distinguishable only when the falling light direction is known. Fig.5 shows that front side dots seems like reverse side dots and vice versa if the image is 180º rotated. We avoided this ambiguity by taking all images in light falling from the approximately top or top-left side of the page (Fig.5).

The dataset is divided into 191 training (80%) and 49 test (20%) images. Also, 44 images of various non-Braille texts from the Internet were added to the training set as negative examples.

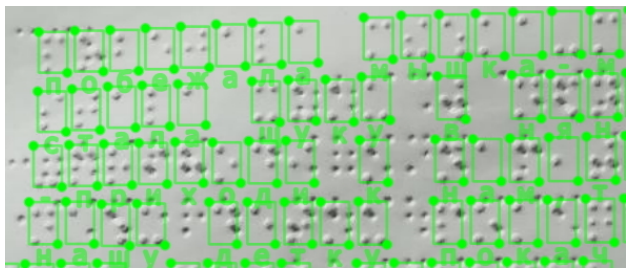

Fig.4    Sample annotations of Angelina Braille Images Dataset.

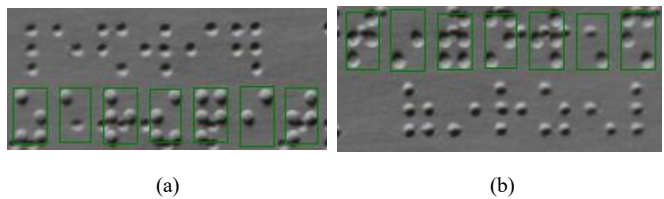

Fig.5    Front side bulging dots (inside green rectangles) looks like reverse side bulged-in dots and vice versa if the image (a) is 180º rotated (b).

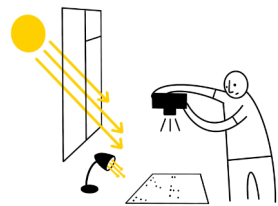

Fig.6    Light conditions Angelina Braille Images Dataset images were taken.

Braille characters classes are labeled by corresponding Russian plain text letters and symbols, but these labels can be converted to braille dots using software tools provided.

Angelina Braille Images Dataset is available at https://github.com/IlyaOvodov/AngelinaDataset.

## IV    EXPERIMENTS SETUP AND RESULTS

### A   Metrics used

We evaluated precision, recall and F1 metrics at character and pixel level. Since the algorithm immediately detects symbols, it is natural for it to determine the precision at the symbol level. The calculation method used by us coincides with [14]. Detected Braille characters that intersect with ground truth characters with IOU (intersection over union) ≥ 0.5 and have a correct class are considered as true positives (*TP*). Otherwise they are considered as false positive (*FP*). So *FP* detections include both detections that has incorrect position as well as detections with correct position but incorrect Braille characters assigned to them. Ground truth symbols that does not have corresponding *TP* detected Braille symbols are considered as false negative (*FN*). The *Precision*, *Recall* and *F1* metrics are defined as:

$$Precision = TP / (TP+FP)$$

$$Recall = TP / (TP+FN) \qquad (2)$$

$$F1 = 2 \cdot Precision \cdot Recall / ( Precision + Recall )$$

We also evaluated per-dot metrics to compare our results with known Braille recognition methods based on dot detection. Since our algorithm does not detect individual points, we do it indirectly. All points of *TP* characters are considered as *TP* points. All points of ground truth characters that do hot have detection intersecting with them with IOU≥0.5 are considered *FN* points, and all points of detection characters that do not intersect with ground truth with IOU≥0.5 are *FP* points. For detections that intersect with ground truth characters with IOU≥05 but has character other than g.t.

character, we compare the presence of point at each of 6 places of the g.t. and detected Braille characters. If some point present in both characters, it is considered as *TP*. Otherwise, it is considered as *FP* and *FN*, respectively. Then we evaluate Precision, Recall, and F1 at the dot level in a way described above (2).

*B  Network training*

We used the DSBI dataset [6] to compare the effectiveness of our algorithm with approaches published earlier ([6], [14]). Train-test split defined for DSBI dataset contains only 28 train images, which is too small for CNN training. We defined a train set as the first 74% and a test set as the last 26% pages of each Braille book in the DSBI dataset. It resulted in 84 and 30 images for train and test sets, respectively.

To evaluate our algorithm in more complex conditions, we combined the train sets from the DSBI dataset described above and our Angelina Braile Images Dataset. The evaluation was performed separately on the DSBI and Angelina test sets.

We trained the neural network to handle images resized to 864-pixel width, which corresponds to approximately 100dpi. When training the neural network, to obtain a better resistance to different image scales and possible input distortions, each input image was scaled to a random width from 550 to 1150pix, which is ± 30% of the required width. The resulting image height was determined so that the image was compressed or stretched vertically by a random amount within ± 10%. Then, we rotated images at a random angle within ± 5 degrees. With a probability of 50%, we reflected the image along the vertical axis and changed each character label to the reflected character's label.

We normalized the image using formula

$$x_c = \frac{I_c - m}{3 \max{(s, 0.1*255)}} \quad (3)$$

where $I_c$ is the intensity of image channel *c*, *m* is the mean of $I_c$ over the whole image and *s* is the standard deviation of $I_c$ over the image.

The random 416x416 image crop was used as CNN input. We trained the neural network for 500 epochs using Adam optimizer [24] with learning rate = 1e-4 and batch size = 24. Initially the $\lambda_{cls}$ factor in the loss function (1) was set to 1. In this case, $L_{loc}$ component of the loss function prevails on $L_{cls}$ resulting in more fast learning of character position than character classification. After 500 epochs we set $\lambda_{cls}$ to 100 making contribution of both components approximately similar. We noticed that if contribution of both loss function components is set equal from the beginning, the learning process becomes unstable. Finally, we set $\lambda_{cls} = 1000$ and train the CNN using the "Reduce On Plateau" approach, i.e., reducing learning rate by factor 10 if the F1 metric on the test set does not decrease for 500 epochs.

*C  Results and discussion*

Table I shows results of experiments.

When trained on the DSBI dataset, the new method gives F1=0.9976 in a character-based test and F1=0.9994 in a dot-based test. It outperforms other methods.

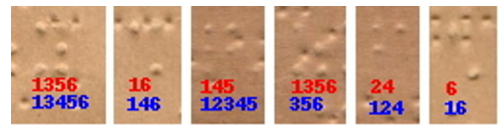

Fig.7  Some characters in DSBI dataset being counted as errors. Red numbers – ground truth Braille character dots, blue – detected Braille character dots. It is not clear, what one is correct.

Investigation of characters that cause errors shows that in many cases, the correct label is questionable (Fig.7) The decision of whether detection or ground truth label is correct is, to a significant extent, subjective. Therefore a further comparison of algorithms with F1≥0.997 on the DSBI dataset seems to be not informative. The reason is that the DSBI dataset was labeled without considering the semantic meanings of characters. So ambiguous cases were labeled arbitrary. The Angelina Dataset proposed in this paper was labeled using Braille characters' semantic meaning, so questionable cases were resolved in favor of an option that should present at the specific place from the grammatical point of view.

When trained on both DSBI and Angelina Braille Images Dataset, our method results in F1=0.9981 on the Angelina test set and F1=0.9963 on DSBI. One can assume that some decrease of precision on this relatively uniform specific dataset is caused by the fact that CNN was trained on more diversified data that requires more generalization ability. As we can see, paper deformations and perspective distortion that present in photos in this dataset do not make recognition quality worse than on the DSBI dataset, where pages are flattened and scanned without distortion.

Processing of one A4 page with proposed method takes time 0.18 s/image on GPU NVIDIA 1080Ti. It outperforms other methods, including BraUNet [14], for which 0.25s/image is reported for the same hardware.

CONCLUSION

The proposed Braille recognition algorithm has shown high accuracy and performance. The algorithm is resistant to irregularities and perspective distortions of the depicted sheet with text, making it possible to recognize texts captured with a mobile phone in everyday conditions.

Angelina Braille Images Dataset was created and published to train the neural network and evaluate the quality of algorithms on images with a large distortion. The dataset is available at https://github.com/IlyaOvodov/AngelinaDataset.

Source code of the algorithm is available at https://github.com/IlyaOvodov/AngelinaReader.

Web service Angelina Braille Reader for recognition of images with Braille texts is available at http://angelina-reader.ru.

ACKNOWLEDGMENT

I would like to express my deep gratitude to the creators of the DSBI dataset [6] for publishing it. Without it, this work would have been impossible.

TABLE I. EXPERIMENTSL RESULTS.

| Method | Train dataset | Test dataset | Braille dot level | | | Character level | | | Performance, s/image |
|---|---|---|---|---|---|---|---|---|---|
| | | | *Precision* | *Recall* | *F1* | *Precision* | *Recall* | *F1* | |
| Segment [6] | DSBI [6] | DSBI | 0.9172 | 0.9811 | 0.948 | | | | |
| Haar [6] | | | 0.9765 | 0.9638 | 0.970 | | | | |
| Haar [10] | | | 0.9838 | 0.9575 | 0.970 | | | | 0.89 |
| HOG LBP SVM [10] | | | 0.9314 | 0.9869 | 0.958 | | | | 15.02 |
| SVM Grid [10] | | | 0.9931 | 0.9997 | 0.996 | | | | 1.22 |
| TS-OBR [10] | | | 0.9965 | 0.9997 | 0.998 | 0.9928 | 0.9996 | 0.9962 | 1.45 |
| BraUNet [14] | | | | | | 0.9943 | 0.9988 | 0.9966 | 0.25 |
| Our | DSBI | DSBI | 0.9992 | 0.9995 | **0.9994** | 0.9977 | 0.9975 | **0.9976** | 0.18 |
| | DSBI + Angelina (our) | | 0.9984 | 0.9993 | 0.9989 | 0.9961 | 0.9964 | 0.9963 | 0.18 |
| | DSBI | Angelina (our) | 0.9812 | 0.9143 | 0.9466 | 0.9569 | 0.898 | 0.9265 | 0.18 |
| | DSBI + Angelina (our) | | 0.9995 | 0.9986 | **0.9991** | 0.9985 | 0.9978 | **0.9981** | 0.18 |